\documentclass{article}
\usepackage{spconf,amsmath,graphicx}

\usepackage{graphicx}
\usepackage{booktabs}
\usepackage{multirow}
\usepackage{algorithm}
\usepackage{amssymb}
\usepackage{pifont}
\usepackage{balance}
\usepackage{color}
\usepackage{amsmath}
\usepackage{lipsum}
\newcommand{\figref}[1]{Fig.~\ref{#1}}

\newcommand{\tableref}[1]{Table~\ref{#1}}
\usepackage{xspace}
\makeatletter
\DeclareRobustCommand\onedot{\futurelet\@let@token\@onedot}
\def\@onedot{\ifx\@let@token.\else.\null\fi\xspace}
\def\eg{\emph{e.g}\onedot} 
\def\ie{\emph{i.e}\onedot} 
 
\def\etc{\emph{etc}\onedot} 
 
\def\etal{\emph{et al}\onedot}

\title{Spy-Watermark: Robust Invisible Watermarking for Backdoor Attack}

\name{Ruofei Wang\textsuperscript{\rm 1}, Renjie Wan\textsuperscript{\rm 2}, Zongyu Guo\textsuperscript{\rm 1}, Qing Guo\textsuperscript{\rm 3,4}, Rui Huang\textsuperscript{\rm 1,*} \thanks{*Rui Huang is the corresponding author: rhuang@cauc.edu.cn.}}
\address{
\textsuperscript{\rm 1}College of Computer Science and Technology, Civil Aviation University of China \\
\textsuperscript{\rm 2}Department of Computer Science, Hong Kong Baptist University\\
\textsuperscript{\rm 3}IHPC, Agency for Science, Technology and Research, Singapore\\
\textsuperscript{\rm 4}CFAR, Agency for Science, Technology and Research, Singapore
}
\let\OLDthebibliography\thebibliography
\renewcommand\thebibliography[1]{
  \OLDthebibliography{#1}
  \setlength{\parskip}{0pt}
  \setlength{\itemsep}{3pt plus 0.3ex}
}

\begin{document}

\setlength{\lineskiplimit}{0pt}
\setlength{\lineskip}{0pt}
\setlength{\abovedisplayskip}{1pt}
\setlength{\belowdisplayskip}{1pt}
\setlength{\abovedisplayshortskip}{1pt}
\setlength{\belowdisplayshortskip}{1pt}

%
\maketitle
\begin{abstract}
Backdoor attack aims to deceive a victim model when facing backdoor instances while maintaining its performance on benign data. Current methods use manual patterns or special perturbations as triggers, while they often overlook the robustness against data corruption, making backdoor attacks easy to defend in practice. To address this issue, we propose a novel backdoor attack method named \textit{Spy-Watermark}, which remains effective when facing data collapse and backdoor defense. Therein, we introduce a learnable watermark embedded in the latent domain of images, serving as the trigger. Then, we search for a watermark that can withstand collapse during image decoding, cooperating with several anti-collapse operations to further enhance the resilience of our trigger against data corruption. Extensive experiments are conducted on CIFAR10, GTSRB, and ImageNet datasets, demonstrating that \textit{Spy-Watermark} overtakes ten state-of-the-art methods in terms of robustness and stealthiness.
\end{abstract}

\section{Introduction}
\label{sec:intro}

Backdoor attack injects triggers into images, misleading the network to output given labels for backdoor data yet retains performance on clean data \cite{gu2017badnets}. Backdoor attack has received extensive attention due to its crucial role in secure and robust deep learning, especially in face recognition~\cite{2017Targeted}, speaker verification~\cite{zhai2021backdoor} autonomous driving~\cite{2020Adversarial}, and medical diagnosis~\cite{2019Adversarial}. 
The most two significant concerns for a backdoor attack model are the \textit{stealthiness of triggers} and \textit{attack success rate} \cite{guo2022overview}. A great stealthy trigger can fool humans and backdoor defense methods to attack the victim model and hardly sacrificing the clean data accuracy. 


%
\begin{figure}[t]
    \centering
    \includegraphics[width=\linewidth]{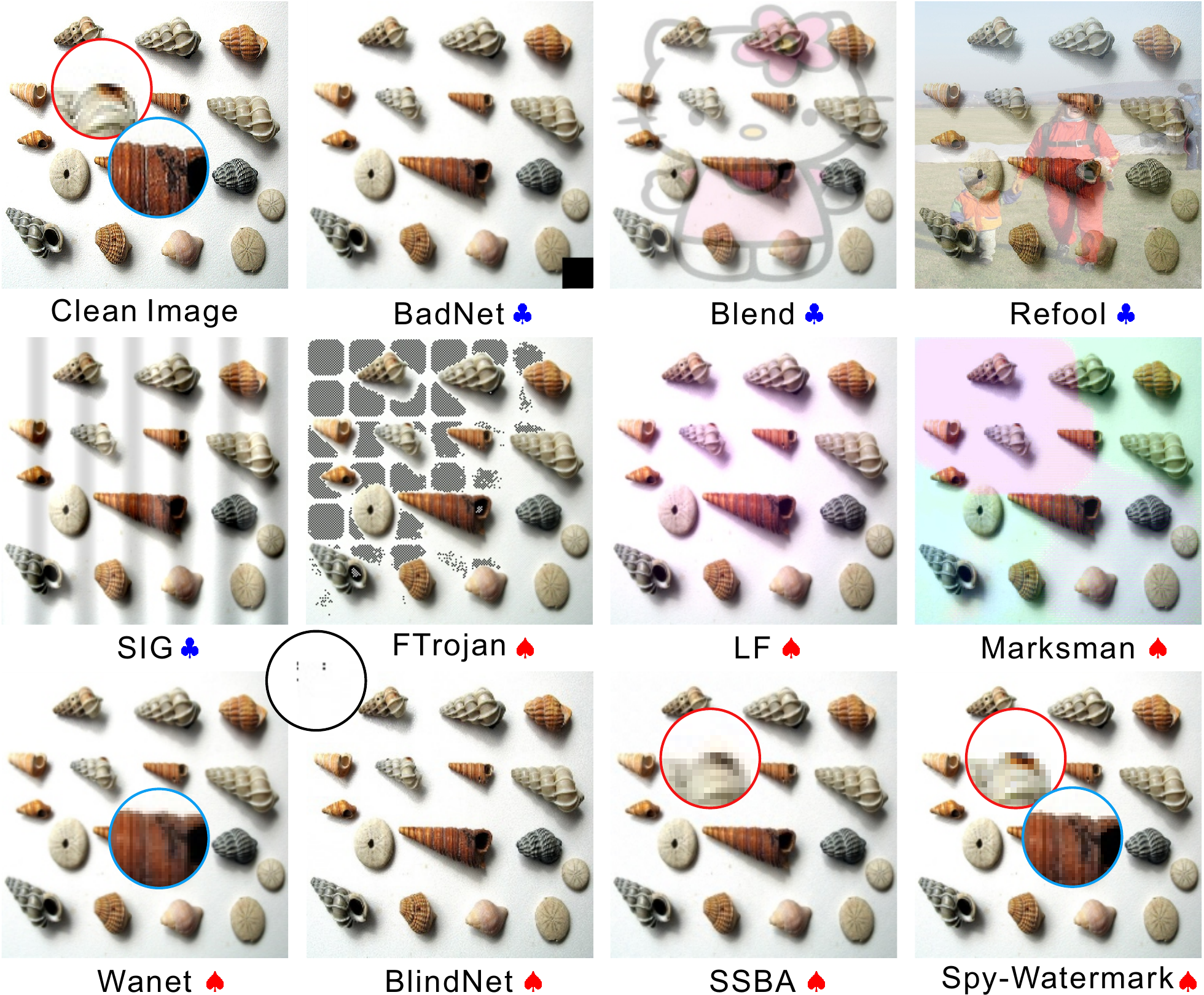}
    \caption{Poisoned examples on the ImageNet dataset. Challenging regions are zoomed in the circle for a clear view. 
    {\color{blue}$\clubsuit$} and {\color{red}$\spadesuit$} denote visible and invisible triggers, respectively.}
    \label{fig:motivate}
   
\end{figure}
\begin{figure*}[ht]
    \centering
\includegraphics[width=0.9\linewidth]{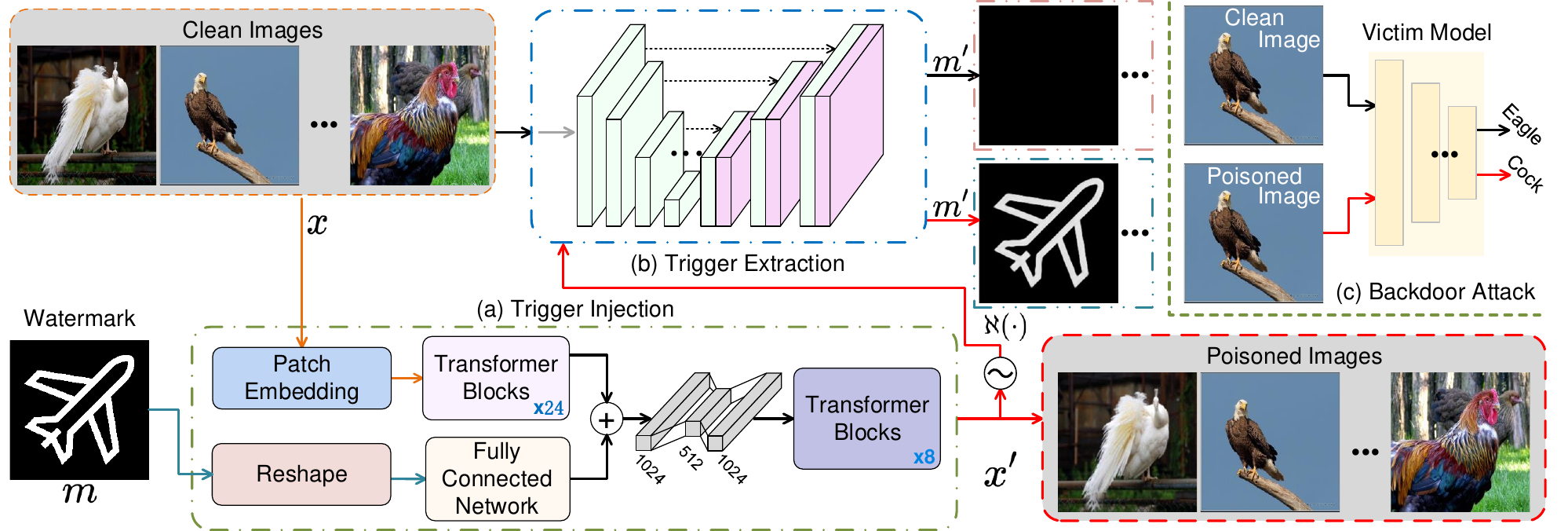}
    \caption{The overall pipeline of our \textit{Spy-Watermark} includes trigger injection, trigger extraction, and backdoor attack.}
    \label{fig:framework}
    \vspace{-5pt}
\end{figure*}
%


To achieve better stealthiness, most of the recent backdoor attack methods are prone to use invisible triggers.  
Li et al. \cite{li2020invisible} propose to use steganography to generate invisible backdoor triggers by replacing some information in a given pixel. 
Zhong \etal. \cite{zhong2020backdoor} treat the pixel intensity change as the invisible trigger. 
Li \etal \cite{li2021invisible} propose extracting the sample-specific perturbation from each training image to poison the data.
Feng \etal \cite{feng2022fiba} create poisoned images by combining the triggers and original images in the frequency domain.
Zhang \etal \cite{zhang2022poison} present injecting the object edges with different colors to poison the data, named `poison ink'.
Although these methods can achieve a relatively high success rate in attacking, they would easily be destroyed by data corruption or cleansed by humans (see \figref{fig:motivate}). 
Obviously the more robust trigger is, the easier to get a high attack success rate.
Therefore, we mainly focus on the robustness of triggers.


In this paper, we propose a novel framework, \ie, \textit{Spy-Watermark}, to embed a learnable watermark into the latent domain of images as the trigger to attack the victim model. \textit{Spy-Watermark} consists of three parts: trigger injection, extraction, and backdoor attack. Firstly, we design a transformer-based trigger injection module to embed the watermark in a learning-based strategy. Next, the trigger extraction network is designed to extract the embedded watermark from the poisoned images, ensuring the presence of the injected trigger against data corruption. To further improve the robustness of our trigger, a set of trigger anti-collapse operations is proposed to force the injection module can inject the watermark under different malicious conditions successfully. Finally, extensive experiments conducted on CIFAR10, GTSRB, and ImageNet datasets demonstrate that our approach has a higher attack success rate and clean data accuracy than the SOTA backdoor attackers.


\section{Method}
\subsection{Backdoor attack formulation}
Let $\mathcal{T}=\{<x_i, y_i>\}^{N}_{i=1}$ denotes the training image and label set, $f_{\theta}$ represents a classification network with pretrained parameters $\theta$, $t$ is a targeted label. Backdoor attack aims to inject a trigger into $f_{\theta}$ to learn a different mapping relationship between benign images and poisoned $x_i$, as:
\begin{equation}
    f_\theta(x_i) \mapsto y_i, f_\theta(\mathcal{B}(x_i)) \mapsto t,
\end{equation}
where $\mathcal{B}(\cdot)$ denotes the trigger injection function, which fuses an image with a backdoor trigger $m$ by $\lambda$:
\begin{equation}
    \mathcal{B}(x)=x\times(1-\lambda) + m\times\lambda.
\end{equation}
$\lambda$ controls the stealthiness and robustness.
We formulate trigger injection operation $\mathcal{B}$ as image watermarking to improve the stealthiness of the injected trigger. \figref{fig:framework} shows the proposed pipeline of \textit{Spy-Watermark}, which consists of trigger injection, trigger extraction, and backdoor attack.


\begin{table*}[!h]
    \centering
    \caption{Quantitative results of the SOTA attackers under random masking~(RM), rotation~(Ro), noising~(Noise) and re-scaling~(RS) on CIFAR10, GTSRB and ImageNet datasets. The \textbf{bold} is the best.}
    \resizebox{0.9\linewidth}{!}{
    \begin{tabular}{llcccccccccccc}
    \toprule
    \multirow{2}{*}{}&\multirow{2}{*}{Method}&\multicolumn{6}{c}{CDA\%}&\multicolumn{6}{c}{ASR\%} \\
    \cmidrule(r){3-8}\cmidrule(r){9-14}
    
    &&None&RM&Ro&Noise&RS&AVG&None&RM&Ro&Noise&RS&AVG\\
 
    \midrule
    \multirow{11}{*}{\rotatebox{90}{CIFAR10}}&BadNet~\cite{gu2017badnets} & 91.5  & 13.2 & 72.7 & 89.2 & 84.8 &70.3  & 96.2 & 74.5 & 33.9 & 95.5 & 94.5 &78.9\\
    
    &Blend~\cite{2017Targeted} & 93.4 & 16.1 & 79.7 & 92.7 & 87.5 &73.9  & 99.9 & 0.0 & 49.3 & 99.9 & 66.5 & 63.1\\
    
    &Refool~\cite{liu2020reflection} & 77.0 & 25.5 & 52.4 & 76.6 & 75.6 &61.4 & 87.0 & 5.8 &19.6 & 86.8 & 85.0&56.8 \\

    &BlindNet~\cite{kwon2022blindnet} &85.4&27.0&51.4&85.0&65.8&62.9  &88.6&41.6&30.6&88.8&36.6& 57.2\\
    
    &SSBA~\cite{li2021invisible} &92.8 & 10.1 & 78.1 & 91.0 & 86.8 & 71.8 & 98.6 & 99.0 & 60.2 & 98.8 & 67.5 & 84.8\\

    &SIG~\cite{barni2019new} &  84.5 & 15.4 & 72.3 & 84.2 & 79.0 &67.1  & 98.6 & 15.6 & 54.5 & 97.1 & 96.5 &72.5\\
    
     &FTrojan~\cite{wang2021backdoor} & 93.5 & 13.3 & 79.5 & 92.4 & 87.8 &73.3  & \textbf{100} & \textbf{100} & 16.8  & \textbf{100} & 59.1 &75.2\\
        
    &LF~\cite{zeng2021rethinking} & 77.0 & 11.8 & 64.6 & 76.6 & 70.2 &60.0 & 95.9 & 2.8 & 95.8 & 95.5 & 95.4 & 77.1\\
        
    &WaNet~\cite{nguyen2020wanet} & 91.6 & 12.9 & 77.7 & 91.1  & 83.5 & 71.4 & 85.6 & 66.0 & 87.8 & 85.0 & 87.4 & 82.4\\
    
    &Marksman~\cite{doanmarksman}&58.1&47.9&49.8&58.5&55.1&53.9 &99.7&99.6&99.7&99.7&99.7&\textbf{99.7}\\
    
    &Spy-Watermark &\textbf{94.7}&\textbf{51.6}&\textbf{93.0}&\textbf{94.4}&\textbf{94.2}&\textbf{85.6}&\textbf{100} &34.1&\textbf{100}&99.9&\textbf{100}&86.8\\

    
    
    
    

    
    


    
    
    \midrule
    \multirow{11}{*}{\rotatebox{90}{GTSRB}}&BadNet~\cite{gu2017badnets} & 96.7  & 8.8 & 82.3 & 96.8 & 96.5 & 76.2 &  92.7  & 95.3 & 14.8  & 92.6 & 50.0 & 69.1 \\
    
    &Blend~\cite{2017Targeted} & 98.4 & 15.7  & 87.0 & 98.3 & 98.1 & 79.5  & \textbf{100} & 99.7 & 96.3 & \textbf{100} & 99.3 & 99.1\\
    
    &Refool~\cite{liu2020reflection} & 98.6 & \textbf{61.6} & 78.2 &  98.5 & \textbf{98.3} & \textbf{87.0} & 62.9 & 43.8 & 25.7 & 
    62.9 & 58.6 & 50.8 \\
    
    &BlindNet~\cite{kwon2022blindnet}&\textbf{100}&4.4&75.7&\textbf{98.8}&92.2&74.2 &91.5&95.4&18.9&91.7&23.2&64.1 \\
    
    &SSBA~\cite{li2021invisible} &98.2 & 0.5 & 86.0 & 98.2 & 97.9 & 76.2  & \textbf{100} & \textbf{100} & 99.5 & \textbf{100} & 97.6 & 99.4\\
    
    &SIG~\cite{barni2019new} & 98.4  & 28.1 & 86.5 & 98.4 & 98.1 & 81.9 & 68.8  & 70.4 & 43.2 & 68.3 & 67.5 & 63.6 \\

    &FTrojan~\cite{wang2021backdoor} & 98.2& 7.4 & 86.4 & 98.2 & 97.9 & 77.6 & \textbf{100}& \textbf{100} & 14.9 &\textbf{100}& 51.6 & 73.3 \\
        
    &LF~\cite{zeng2021rethinking} & 98.2 & 35.5 & 86.5 & 98.2 & 97.8 & 83.2 & \textbf{100} & 99.9 & \textbf{100} & \textbf{100} & \textbf{100} & \textbf{100}\\
    
    &WaNet~\cite{nguyen2020wanet} & 98.7 & 6.0 & 89.5 & 98.5 & 96.7 & 77.9 & 98.1 & 1.3 &  88.8 & 93.1 & 96.7 & 75.6 \\
    &Marksman~\cite{doanmarksman}&50.3&44.1&46.6&54.4&52.9&49.7&98.8&99.1&99.0&99.1&99.0&99.0\\
    
    &Spy-Watermark &\textbf{100}&43.1&\textbf{96.8}&97.1&97.2&86.8&\textbf{100} &\textbf{100}&\textbf{100}&\textbf{100}&\textbf{100} &\textbf{100}\\
    
    \midrule
    \multirow{9}{*}{\rotatebox{90}{ImageNet}}&BadNet~\cite{gu2017badnets} & \textbf{94.0} &  63.8 & 88.2 & 92.4 & \textbf{93.2} &86.3  & \textbf{100}  & 4.0 & 40.0  & 4.0 & \textbf{100}&49.6  \\
    
    &Blend~\cite{2017Targeted} & 91.8  & 60.4  & 87.8 &90.6 &92.0 &84.5 & 98.8 & 42.0& 96.0 &46.0 &\textbf{100} &76.6 \\
    
    &Refool~\cite{liu2020reflection} &80.0 &58.0 &67.9 &78.6 &79.6&72.8 &80.0 &40.0 &29.1 &80.0 &80.0&61.8  \\
    &BlindNet~\cite{kwon2022blindnet}&97.0&69.0&\textbf{94.4}&\textbf{97.0}&92.6&\textbf{90.0} &13.6&15.9&11.6&13.6&11.5&13.2 \\
    
    &SSBA~\cite{li2021invisible}  & 85.2 & 49.8 & 83.2 &78.2 & 86.6 &76.6 & 68.0 & \textbf{100} & 76.0 & \textbf{100}& 32.0 &75.2 \\

    &SIG~\cite{barni2019new}  &90.2 & 60.2 & 88.0 &89.2 & 89.4 &83.4 & 12.0 & 8.0 & 6.0 &6.0 & 6.0 & 7.6\\
    
    &FTrojan~\cite{wang2021backdoor}  & 89.8 &61.2 & 85.2 &89.2 & 90.0 &83.1 &\textbf{100} & 84.0 & 14.0 &84.0 & 44.0 &65.2 \\
    
    &LF~\cite{zeng2021rethinking}  &91.2 &63.6 &87.6 &90.2 &90.2 &84.6 &25.1 &24.6 &25.1 &24.4&25.1 & 24.9\\
     
    &WaNet~\cite{nguyen2020wanet}  & 86.6 &56.4 &86.0 & 86.0&85.4 &80.1 & 86.7 &\textbf{100} &\textbf{100} &84.4&86.7 &91.6 \\
        
    &Marksman~\cite{doanmarksman}&38.0&36.2&39.6&39.0&41.2&38.8 &76.6&76.6&75.4&76.2&76.0&76.2\\
    
    &Spy-Watermark  & 93.2 &\textbf{77.2} &91.2 &91.8 &92.8 &89.2 &\textbf{100} &\textbf{100} &\textbf{100} &\textbf{100}&\textbf{100} &\textbf{100} \\

    \bottomrule
    \end{tabular}
    }
    \vspace{-5pt}
    \label{tab:attack_results}
\end{table*}

\vspace{-5pt}
\subsection{Trigger injection}
To ensure the invisibility and robustness of injected triggers, we have designed a trigger injection module with the Transformer blocks in \figref{fig:framework}(a). 
Injecting a trigger mainly consists of two steps: embedding the trigger into benign images and reconstructing the poisoned counterpart.

Firstly, we use the patch embedding function $P(\cdot)$ to split benign image $x\in\mathbb{R}^{C \times H \times W}$ and then employ the Transformer encoder $Tran^{Enc}_{\times24}(\cdot)$ to extract the corresponding high-level representation. Due to the trigger $m\in\mathbb{R}^{1 \times H \times W}$, we directly use the reshape operation $R(\cdot)$ to compress dimensions and adjust the output channels with a fully connected network $F_{l:1}(\cdot)$. Then, we adopt the addition for fusing the features of $x$ and $m$ together to achieve the goal of trigger embedding. Finally, a lightweight Transformer decoder is used to reconstruct corresponding images from the mixed features. However, using point addition cannot embed the trigger $m$ into the global domain of image features. So we built a three-layer fully connected network $F_{l:3}(\cdot)$ to project trigger features into each neuron. The entire injection process can be formulated as:
\begin{equation}
    x' = Tran^{Dec}_{\times8}(F_{l:3}(Tran^{Enc}_{\times24}(P(x))+F_{l:1}(R(m)))),
\end{equation}
where the number in parentheses means the number of Transformer blocks and fully connected layers. The aforementioned $C, H, W$ are the channel number, height, and width of corresponding images. Lastly, we train the trigger injector by minimizing the following loss:
\begin{equation}
    \mathcal{L}_1 = \frac{1}{HW} \sum_{i,\  j}^{H,W}(F.relu(|x'_{i,\ j}-x_{i,\ j}|-\epsilon)^2),
\end{equation}
where $\epsilon$ is a balance factor for achieving the trade-off between image quality and trigger patterns, which is set to 1/255.
\vspace{-10pt}
\subsection{Trigger extraction}
We build the trigger extractor with an Unet-like network $\Phi(\cdot)$ to guarantee that the poisoned $x'$ truly contains the trigger $m$. $\Phi(\cdot)$ takes $x'$ as input and outputs a watermark $m'$ while not output anything meeting clean images. And the extracted $m'$ from triggered image $x'$ should be identical to the $m$. To alleviate wrong extractions that $m' = \Psi(x)$, we force the extractor to generate a \textbf{ZERO} map with a clean image $x$, \ie, $\textbf{0} = \Phi(x)$. We adopt multiscale supervision on the features of $\Psi(\cdot)$, $m'_l = \Phi(x')^l$, where $l$ means the feature extracted from the $l$-th layer of the trigger extractor. The loss to supervise $\Psi(\cdot)$ is defined as:

\begin{equation}
    \mathcal{L}_2 = \sum_{l=1}^{3}\sum_{i,j}^{H, W}\frac{(\psi(\Phi(x'))^l_{i,j}-m_{i,j})^2+(\psi(\Phi(x))^l_{i,j})^2}{HW}
\end{equation}
where $\psi(\cdot)$ denotes the bilinear upsampling. We joined the trigger injector and extractor together to train in an end-to-end way, so the final loss $\mathcal{L}=\lambda_1 \mathcal{L}_1+\lambda_2 \mathcal{L}_2$. The balance weights $\lambda_1$ and $\lambda_2$ are set to be 1.0 and 0.1, respectively.
\vspace{-5pt}
\subsection{Trigger anti-collapse}
\label{sec:bdc}
Due to the robustness of the injected trigger affecting the performance of backdoor models, we propose a trigger anti-collapse operation set $\mathbb{S}=\{s_1, s_2, ..., s_n\}$ to increase the robustness of our trigger against different data corruptions, where $s_k$ indicate different image corrupting operations, \ie random masking, re-scaling, noising, rotation, \etc. For example, random masking is implemented to erase the 1/4 areas of poisoned images, which pushes our injector to embed the trigger into the global area of sampled images. And the re-scaling is to adjust the image scale from $0.5\times$ to $2\times$ of the original resolution, which would facilitate the retention of our trigger after data compression and transmission. Noising and rotation are designed to meet diverse input requirements. The anti-collapse process can be formulated as:
\begin{equation}
    x'= \aleph_k^n(x', s_k, o_k, p_k)
\end{equation}
where $o_k$ denotes the operation-related parameters and $p_k$ represents the probability of adopting $o_k$ to attack $x'$. By taking this strategy, minimizing the loss of the extractor will motivate our injector to embed $m$ into images robustly.



\section{Experiment}
\label{sec:exper}

\subsection{Setup}

\textbf{Dataset.} Our experiments are conducted on three public datasets including CIFAR10 \cite{krizhevsky2009learning}, GTSRB \cite{stallkamp2011german} and ImageNet \cite{deng2009imagenet}. 
Due to computational limitations, we randomly select 10 classes from the ImageNet dataset to evaluate \textit{Spy-Watermark} and the comparison methods. 


\textbf{Evaluation Metrics.} We use PSNR, SSIM, and LPIPS to evaluate the invisibility of different trigger patterns. For the backdoor attack, we use Clean Data Accuracy (CDA) and Attack Success Rate (ASR) to compare the performance of various attackers.

\textbf{Implementation Details.}
For trigger injection, we train the trigger injector for 10k iterations by SGD optimizer with the learning rate of 2e-4, momentum of 0.5 and batch size of 16. 
For backdoor attack, the poison ratio~($\rho$) is 0.1 and the first category of each dataset is set as the target label.
ResNet18 \cite{he2016deep} is used as the victim model. We optimize the network by SGD with a momentum of 0.9, learning rate of 1e-1, and epoch of 100. And the learning rate will decay every epoch by the Cosine Annealing. 
Experiments are implemented with Pytorch~\cite{paszke2019pytorch} and conducted on a NVIDIA RTX 3090 GPU.
%
\subsection{Evaluation of attack}
\label{sec:attack_experiment}

\tableref{tab:attack_results} shows the quantitative results of our \textit{Spy-Watermark} and 10 SOTA backdoor attackers on three public datasets. 
On CIFAR10, compared with Blend and FTrojan, our attacker achieves 11.7\% and 12.3\% of relative CDA improvements, respectively. Marksman achieves the highest ASR by 99.7\%, which is bought by the sacrifice of the CDA. \textit{Spy-Watermark} achieves higher CDA than it by 31.7\%. 
We further evaluate the effectiveness of \textit{Spy-Watermark} on a traffic sign recognition dataset, \ie GTSRB. Compared with the frequency domain attack model, LF, \textit{Spy-Watermark} improves the average CDA from 83.2\% to 86.8\%. BlindNet injects triggers in the frequency domain by Fourier Transform, \textit{Spy-Watermark} gets 12.6\% and 35.9\% relative improvements in terms of average metrics on clean and poisoned data. 
%
On the ImageNet dataset, compared with the BadNet, our model improves the average CDA and ASR by 2.9\% and 50.4\%, respectively. As for comparing with image-warping based attacker-WaNet, \textit{Spy-Watermark} promotes the average CDA from 80.1\% to 89.2 and ASR from 91.6\% to 100.0\%. 
In total, the proposed \textit{Spy-Watermark} is more robust against different defense operations than other backdoor attackers.

    

    

\subsection{Backdoor defense}
\label{sec:defen}
\begin{figure}[t]
    \centering
    \includegraphics[width=\linewidth]{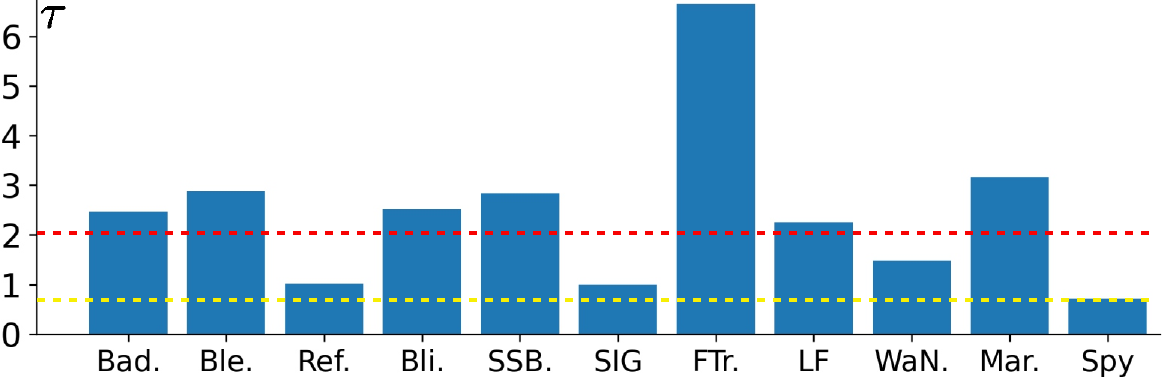}
    \caption{NC backdoor defense results of each backdoor method tested on CIFAR10 dataset. The metric $\tau=2$ (red dashed line) donates the threshold for clean and backdoor patterns. Due to the space limitations, we have abbreviated the names of each backdoor method.}
    \label{fig:defense}
    \vspace{-10pt}
   \end{figure}
Neural Cleanse (NC)~\cite{wang2019neural} is a powerful backdoor detection and mitigation method to improve the security of a DNN model. We utilize the NC to defend each aforementioned backdoor method for evaluating the robustness of backdoor models. Only four methods achieve anomaly metrics less than 2 (under the red dashed line in \figref{fig:defense}), indicating the successful avoidance of the backdoor defense. Compared with ten SOTA methods, our \textit{Spy-Watermark} achieves the lowest anomaly metric by 0.7, which demonstrates that \textit{Spy-Watermark} is the most robust backdoor attacker against the backdoor defense.

\begin{table}[h]
\vspace{-10pt}
    \centering
    \caption{Comparison of the invisibility (Stealthiness) of different SOTA backdoor attack models on the CIFAR10 dataset. The \textbf{bold} is the best.}
    \resizebox{1\linewidth}{!}{
    \Huge 
    \begin{tabular}{lccccccccccc} 
    \toprule
    &Bad.&Ble.&Ref.&Bli.&SSB. &SIG&FTr. & LF & WaN.&Mar.&Spy.\\
    \midrule
    PSNR$\uparrow$ & 27.23 &20.75 &15.26 &34.29&24.83 &19.32 &33.78 &23.03 &23.23 &24.56&\textbf{38.73}\\
    SSIM$\uparrow$ & 0.992 &0.803 &0.574 &0.992&0.900  &0.620& 0.962 &0.963 &0.920 &0.853&\textbf{0.995}\\
    LPIPS$\downarrow$&0.104&0.123&0.117&0.001&0.113 & 0.201 &0.110 &0.123 &0.113 & 0.176&\textbf{0.001}\\

    \bottomrule
    \end{tabular}
    }
    
    \label{tab:mark_cifar10}
    \vspace{-20pt}
\end{table}

\subsection{Evaluation of stealthiness}
\label{sec:steal}

\tableref{tab:mark_cifar10} shows the quantitative results of different trigger injection methods on the CIFAR10 dataset.
The higher the image quality of poisoned data, the better the stealthiness of triggers, and the more difficult defend by backdoor defense methods and humans.
Among the compared methods, \textit{Spy-Watermark} achieves the highest PSNR (38.73),  SSIM (0.995), and the lowest LPIPS (0.001). FTrojan is the second most stealthy method, our method achieves 4.95, 0.033, and 0.109 improvements than it in terms of PSNR, SSIM, and LPIPS.


\section{Conclusion}
\label{sec:conc}
We propose incorporating an invisible watermark into the latent domain of images as the trigger to implement the backdoor attack, \ie \textit{Spy-Watermark}. To ensure the robustness of our injected trigger, the trigger extraction module and a series of trigger anti-collapse operations are introduced to compel the injector to embed the watermark secretly and effectively.
We have conducted abundant experiments on three datasets (\eg, CIFAR10, GTSRB, and ImageNet) to showcase the stealthiness and attack capabilities of \textit{Spy-Watermark}. And also a  typical backdoor defense method is adopted to verify the robustness of our triggers. The experimental results demonstrate the superiority of \textit{Spy-Watermark} over ten state-of-the-art methods. 

\bibliographystyle{IEEEbib}
\bibliography{refs}

\begin{thebibliography}{10}

\bibitem{gu2017badnets}
Tianyu Gu, Brendan Dolan-Gavitt, and Siddharth Garg,
\newblock ``Badnets: Identifying vulnerabilities in the machine learning model
  supply chain,''
\newblock {\em arXiv preprint arXiv:1708.06733}, 2017.

\bibitem{2017Targeted}
Xinyun Chen, Chang Liu, Bo~Li, Kimberly Lu, and Dawn Song,
\newblock ``Targeted backdoor attacks on deep learning systems using data
  poisoning,''
\newblock {\em arXiv preprint arXiv:1712.05526}, 2017.

\bibitem{zhai2021backdoor}
Tongqing Zhai, Yiming Li, Ziqi Zhang, Baoyuan Wu, Yong Jiang, and Shu-Tao Xia,
\newblock ``Backdoor attack against speaker verification,''
\newblock in {\em ICASSP 2021-2021 IEEE International Conference on Acoustics,
  Speech and Signal Processing (ICASSP)}. IEEE, 2021, pp. 2560--2564.

\bibitem{2020Adversarial}
Ranjie Duan, Xingjun Ma, Yisen Wang, James Bailey, A~Kai Qin, and Yun Yang,
\newblock ``Adversarial camouflage: Hiding physical-world attacks with natural
  styles,''
\newblock in {\em Proceedings of the IEEE/CVF CVPR}, 2020, pp. 1000--1008.

\bibitem{2019Adversarial}
S.~G. Finlayson, J.~D. Bowers, J.~Ito, J.~L. Zittrain, Andrew~L. Beam, and
  I.~S. Kohane,
\newblock ``Adversarial attacks on medical machine learning,''
\newblock {\em Science}, vol. 363, no. 6433, pp. 1287--1289, 2019.

\bibitem{guo2022overview}
Wei Guo, Benedetta Tondi, and Mauro Barni,
\newblock ``An overview of backdoor attacks against deep neural networks and
  possible defences,''
\newblock {\em IEEE Open Journal of Signal Processing}, 2022.

\bibitem{li2020invisible}
Shaofeng Li, Minhui Xue, Benjamin Zi~Hao Zhao, Haojin Zhu, and Xinpeng Zhang,
\newblock ``Invisible backdoor attacks on deep neural networks via
  steganography and regularization,''
\newblock {\em IEEE TDSC}, vol. 18, no. 5, pp. 2088--2105, 2020.

\bibitem{zhong2020backdoor}
Haoti Zhong, Cong Liao, Anna~Cinzia Squicciarini, Sencun Zhu, and David Miller,
\newblock ``Backdoor embedding in convolutional neural network models via
  invisible perturbation,''
\newblock in {\em Proceedings of the Tenth ACM CODASPY}, 2020, pp. 97--108.

\bibitem{li2021invisible}
Yuezun Li, Yiming Li, Baoyuan Wu, Longkang Li, Ran He, and Siwei Lyu,
\newblock ``Invisible backdoor attack with sample-specific triggers,''
\newblock in {\em Proceedings of the IEEE/CVF ICCV}, 2021, pp. 16463--16472.

\bibitem{feng2022fiba}
Yu~Feng, Benteng Ma, Jing Zhang, Shanshan Zhao, Yong Xia, and Dacheng Tao,
\newblock ``Fiba: Frequency-injection based backdoor attack in medical image
  analysis,''
\newblock in {\em Proceedings of the IEEE/CVF CVPR}, 2022, pp. 20876--20885.

\bibitem{zhang2022poison}
Jie Zhang, Chen Dongdong, Qidong Huang, Jing Liao, Weiming Zhang, Huamin Feng,
  Gang Hua, and Nenghai Yu,
\newblock ``Poison ink: Robust and invisible backdoor attack,''
\newblock {\em IEEE TIP}, vol. 31, pp. 5691--5705, 2022.

\bibitem{liu2020reflection}
Yunfei Liu, Xingjun Ma, James Bailey, and Feng Lu,
\newblock ``Reflection backdoor: A natural backdoor attack on deep neural
  networks,''
\newblock in {\em ECCV}. Springer, 2020, pp. 182--199.

\bibitem{kwon2022blindnet}
Hyun Kwon and Yongchul Kim,
\newblock ``Blindnet backdoor: Attack on deep neural network using blind
  watermark,''
\newblock {\em Multimedia Tools and Applications}, pp. 1--18, 2022.

\bibitem{barni2019new}
Mauro Barni, Kassem Kallas, and Benedetta Tondi,
\newblock ``A new backdoor attack in cnns by training set corruption without
  label poisoning,''
\newblock in {\em ICIP}. IEEE, 2019, pp. 101--105.

\bibitem{wang2021backdoor}
Tong Wang, Yuan Yao, Feng Xu, Shengwei An, and Ting Wang,
\newblock ``Backdoor attack through frequency domain,''
\newblock {\em arXiv preprint arXiv:2111.10991}, 2021.

\bibitem{zeng2021rethinking}
Yi~Zeng, Won Park, Z~Morley Mao, and Ruoxi Jia,
\newblock ``Rethinking the backdoor attacks' triggers: A frequency
  perspective,''
\newblock in {\em Proceedings of the IEEE/CVF ICCV}, 2021, pp. 16473--16481.

\bibitem{nguyen2020wanet}
Tuan~Anh Nguyen and Anh~Tuan Tran,
\newblock ``Wanet-imperceptible warping-based backdoor attack,''
\newblock in {\em ICLR}, 2020.

\bibitem{doanmarksman}
Khoa~D Doan, Yingjie Lao, and Ping Li,
\newblock ``Marksman backdoor: Backdoor attacks with arbitrary target class,''
\newblock in {\em NIPS}, 2022.

\bibitem{krizhevsky2009learning}
Alex Krizhevsky, Geoffrey Hinton, et~al.,
\newblock ``Learning multiple layers of features from tiny images,''
\newblock {\em Technical report, University of Toronto}, 2009.

\bibitem{stallkamp2011german}
Johannes Stallkamp, Marc Schlipsing, Jan Salmen, and Christian Igel,
\newblock ``The german traffic sign recognition benchmark: a multi-class
  classification competition,''
\newblock in {\em The 2011 IJCNN}. IEEE, 2011, pp. 1453--1460.

\bibitem{deng2009imagenet}
Jia Deng, Wei Dong, Richard Socher, Li-Jia Li, Kai Li, and Li~Fei-Fei,
\newblock ``Imagenet: A large-scale hierarchical image database,''
\newblock in {\em Proceedings of the IEEE/CVF CVPR}. Ieee, 2009, pp. 248--255.

\bibitem{he2016deep}
Kaiming He, Xiangyu Zhang, Shaoqing Ren, and Jian Sun,
\newblock ``Deep residual learning for image recognition,''
\newblock in {\em Proceedings of the IEEE/CVF CVPR}, 2016, pp. 770--778.

\bibitem{paszke2019pytorch}
Adam Paszke, Sam Gross, Francisco Massa, Adam Lerer, James Bradbury, Gregory
  Chanan, Trevor Killeen, Zeming Lin, Natalia Gimelshein, Luca Antiga, et~al.,
\newblock ``Pytorch: An imperative style, high-performance deep learning
  library,''
\newblock {\em NIPS}, vol. 32, 2019.

\bibitem{wang2019neural}
Bolun Wang, Yuanshun Yao, Shawn Shan, Huiying Li, Bimal Viswanath, Haitao
  Zheng, and Ben~Y Zhao,
\newblock ``Neural cleanse: Identifying and mitigating backdoor attacks in
  neural networks,''
\newblock in {\em 2019 IEEE Symposium on Security and Privacy (SP)}. IEEE,
  2019, pp. 707--723.

\end{thebibliography}

\end{document}